\documentclass{svmult_jk} 
\usepackage{epsfig}
\usepackage{graphicx,amssymb,amsbsy,subfigure}
\usepackage{wrapfig}
\usepackage{psfrag}

\newcommand{\be}{\begin{equation}}
\newcommand{\ee}{\end{equation}}
\newcommand{\beq}{\begin{equation}}
\newcommand{\eeq}{\end{equation}}
\newcommand{\bed}{\begin{displaymath}}
\newcommand{\eed}{\end{displaymath}}

\newcommand{\beqa}{\begin{eqnarray}}
\newcommand{\eeqa}{\end{eqnarray}}

\newcommand{\beqann}{\begin{eqnarray*}}
\newcommand{\eeqann}{\end{eqnarray*}}

\newcommand{\bseq}{\begin{subequations}}
\newcommand{\eseq}{\end{subequations}}

\newcommand{\mat}[2]{\left[ \begin{array}{#1} #2 \end{array} \right] } 
\newcommand{\ba}{\begin{array}}
\newcommand{\ea}{\end{array}}

\newcommand{\1}{{\bf 1}}

\newcommand{\negr}[1]{{\bf {#1}}}

\usepackage{subeqn}

\usepackage{mathptmx}       
\usepackage{helvet}         
\usepackage{courier}        
\usepackage{type1cm}        
\usepackage{makeidx}         
\usepackage{graphicx}        
\usepackage{multicol}        
\usepackage[bottom]{footmisc}

\makeindex             

\begin{document}


\title*{Kinetostatic Performance of a Planar Parallel
Mechanism with Variable Actuation}
\author{N.~Rakotomanga, D.~Chablat and S.~Caro}
\institute{$^1$\'Ecole de Technologie Sup\'erieure, Montr\'eal, QC, Canada, \\ \email{novona.rakotomanga.1@ens.etsmtl.ca} \\
$^2$Institut de Recherche en Communications et Cybern\'etique de Nantes, France, \email{\{damien.chablat, stephane.caro\}@irccyn.ec-nantes.fr}}

%
%
\maketitle

\abstract{This paper deals with a new planar parallel mechanism with variable
actuation and its kinetostatic performance. A drawback of parallel
mechanisms is the non homogeneity of kinetostatic performance within
their workspace. The common approach to solve this problem is the
introduction of actuation redundancy, that involves force control
algorithms. Another approach, highlighted in this paper, is to
select the actuated joint in each limb with regard to the pose of
the end-effector. First, the architecture of the mechanism and two
kinetostatic performance indices are described. Then, the actuating
modes of the mechanism are compared.}

\section{Introduction}
A drawback of serial and parallel mechanisms is the inhomogeneity of the
kinetostatic performance within their workspace. For instance,
dexterity, accuracy and stiffness are usually bad in the
neighbourhood of singularities that can appear in the workspace of
such mechanisms. As far as the parallel mechanisms are concerned, their inverse kinematics problem (IKP) has usually many solutions, which correspond to the
\emph{working modes} of the mechanism~\cite{Chablat:1998}.
Nevertheless, it is difficult to come up with a large workspace free
of singularity with a given working mode. Consequently, a trajectory
planning may require a change of the working mode by means of an
alternative trajectory in order to avoid singular configurations. In
such a case, the initial trajectory would not be followed. The
common approach to solve this problem is to introduce actuation
redundancy, that involves force control algorithms~\cite{AlbaGomez:2005}. Another approach is
to use the concept of joint-coupling as proposed by~\cite{Theingin:2007} or to select the actuated joint
in each limb with regard to the pose of the end-effector,~\cite{Arakelian:2007}, as highlighted in this paper. Therefore,
we introduce a planar parallel mechanism with variable actuation,
also known as \emph{variable actuated mechanism}~(VAM). First, the
architecture of the mechanism and two kinetostatic performance
indices are described. Then, the \emph{actuating modes}~(AMs) of the
mechanism are compared based on their kinetostatic performance.

\section{Preliminaries}
This section deals with the kinematic modeling of a new variable
actuated mechanism~(VAM), its singularity analysis, the presentation
of some performance indices and the concept of regular dextrous
workspace.
\subsection{Mechanism architecture}
The concept of VAM was introduced in~\cite{Arakelian:2007,Theingin:2007}. Indeed,
they derived a VAM from the architecture of the 3-RPR planar
parallel manipulator~(PPM) by actuating either the first revolute
joint or the prismatic joint of its limbs.
\begin{figure}[!ht]
  \centering
  \psfrag{A1}[c][c]{$A_1$}
  \psfrag{B1}[c][c]{$B_1$}
  \psfrag{C1}[c][c]{$C_1$}
  \psfrag{D1}[c][c]{$E_1$}
  \psfrag{E1}[c][c]{$D_1$}
  \psfrag{a1}[c][c]{$\alpha_1$}
  \psfrag{d1}[c][c]{$\delta_1$}
  \psfrag{A2}[c][c]{$A_2$}
  \psfrag{B2}[c][c]{$B_2$}
  \psfrag{C2}[c][c]{$C_2$}
  \psfrag{D2}[c][c]{$E_2$}
  \psfrag{E2}[c][c]{$D_2$}
  \psfrag{a2}[c][c]{$\alpha_2$}
  \psfrag{d2}[c][c]{$\delta_2$}
  \psfrag{A3}[c][c]{$A_3$}
  \psfrag{B3}[c][c]{$B_3$}
  \psfrag{C3}[c][c]{$C_3$}
  \psfrag{D3}[c][c]{$E_3$}
  \psfrag{E3}[c][c]{$D_3$}
  \psfrag{a3}[c][c]{$\alpha_3$}
  \psfrag{d3}[c][c]{$\delta_3$}
  \psfrag{th}[c][c]{$\phi$}
  \psfrag{P}[c][c]{$P$}
  \psfrag{x}[c][c]{$x$}
  \psfrag{y}[c][c]{$y$}
  \psfrag{xp}[c][c]{$x^\prime$}
  \psfrag{yp}[c][c]{$y^\prime$}
  \psfrag{Fb}[c][c]{$\mathcal{F}_b$}
  \psfrag{Fp}[c][c]{$\mathcal{F}_p$}
  \includegraphics[width=8cm]{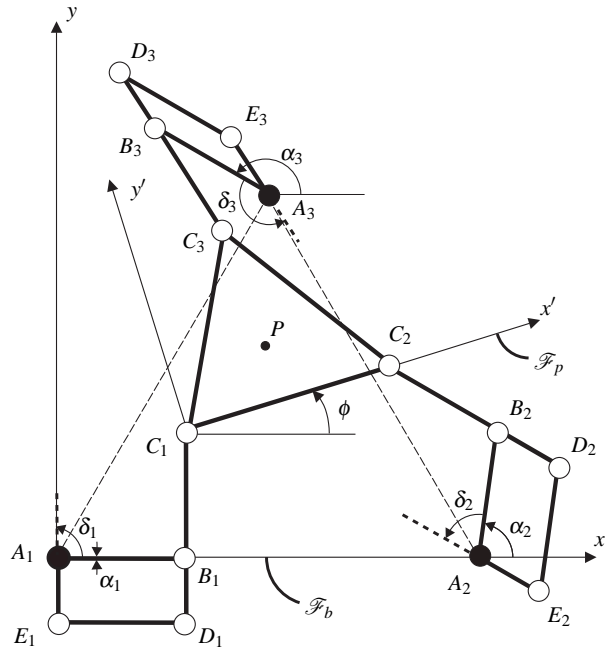}\\
  \caption{3-RRR PPM with variable actuation}
  \label{fig:3RRRvariable}
\end{figure}
This paper deals with the study of a new VAM illustrated in
Figure~\ref{fig:3RRRvariable}. This mechanism is derived from the
architecture of the 3-RRR~PPM. As a matter of fact, the first link
of each limb of the conventional 3-\underline{R}RR manipulator is
replaced by parallelogram $A_iB_iD_iE_i$ to come up with the
mechanism at hand. Accordingly, links $A_iB_i$ and $B_iC_i$ can be
driven independently, i.e., angles $\alpha_i$ and $\delta_i$ are
actuated and uncoupled, by means of an actuator and a double clutch
mounted to the base and located in point $A_i$, $i=1,2,3$.

It turns out that the VAM has eight \emph{actuating modes} as shown
in Table~\ref{tableau_robot}. Indeed, the actuating mode of the
mechanism depends on its actuated joints. For instance, the first
actuating mode corresponds to the 3-\underline{R}RR mechanism, also
called
$\underline{R}RR_1$-$\underline{R}RR_2$-$\underline{R}RR_3$~mechanism
in the scope of this paper, as the first revolute joints (located in
point $A_i$) of its limbs are actuated. Likewise, the eighth
actuating mode corresponds to the 3-R\underline{R}R manipulator,
also called
$R\underline{R}R_1$-$R\underline{R}R_2$-$R\underline{R}R_3$~mechanism,
as the second revolute joints (located in point $B_i$) of its limbs
are actuated.

The moving platform pose of the VAM is determined by means of the
Cartesian coordinates $(x,y)$ of operation point $P$ expressed in
frame $\mathcal{F}_b$ and angle $\phi$, namely, the angle between
frames $\mathcal{F}_b$ and $\mathcal{F}_p$. Moreover, the passive
and actuated joints do not have any stop. Points $A_1$, $A_2$ and
$A_3$, ($C_1$, $C_2$ and $C_3$, respectively) lie at the corners of
an equilateral triangle, of which the geometric center is point $O$
(point $P$, resp.). The length of links $A_iB_i$ and $B_iC_i$ is
equal to~3.0, $i=1,2,3$. The length of segment $A_1A_2$
($C_1C_2$, resp.) is equal to 10.0 (5.0, resp.). The unit is not
specified as absolute lengths are not necessary to convey the idea.

\begin{table}
  \begin{center}
\caption{The eight actuating modes of the 3-RRR VAM}
\begin{tabular}{c|c|c|c}
\hline \hline
\multicolumn{2}{c|}{Actuating mode number} & driven links & active angles \\
\hline 1&$\underline{R}RR_1$-$\underline{R}RR_2$-$\underline{R}RR_3$
& $A_1B_1$, $A_2B_2$, $A_3B_3$ & $\alpha_1$, $\alpha_2$, $\alpha_3$ \\
2&$\underline{R}RR_1$-$\underline{R}RR_2$-$R\underline{R}R_3$ & $A_1B_1$, $A_2B_2$, $A_3E_3$ & $\alpha_1$, $\alpha_2$, $\delta_3$ \\
3&$\underline{R}RR_1$-$R\underline{R}R_2$-$\underline{R}RR_3$ &
$A_1B_1$, $A_2E_2$, $A_3B_3$ & $\alpha_1$, $\delta_2$, $\alpha_3$ \\
4&$R\underline{R}R_1$-$\underline{R}RR_2$-$\underline{R}RR_3$ &
$A_1E_1$, $A_2B_2$, $A_3B_3$ & $\delta_1$, $\alpha_2$, $\alpha_3$ \\
5&$\underline{R}RR_1$-$R\underline{R}R_2$-$R\underline{R}R_3$ &
$A_1B_1$, $A_2E_2$, $A_3E_3$ & $\alpha_1$, $\delta_2$, $\delta_3$ \\
6&$R\underline{R}R_1$-$R\underline{R}R_2$-$\underline{R}RR_3$ &
$A_1E_1$, $A_2E_2$, $A_3B_3$ & $\delta_1$, $\delta_2$, $\alpha_3$ \\
7&$R\underline{R}R_1$-$\underline{R}RR_2$-$R\underline{R}R_3$ &
$A_1E_1$, $A_2B_2$, $A_3E_3$ & $\delta_1$, $\alpha_2$, $\delta_3$ \\
8&$R\underline{R}R_1$-$R\underline{R}R_2$-$R\underline{R}R_3$ &
$A_1E_1$, $A_2E_2$, $A_3E_3$ & $\delta_1$, $\delta_2$, $\delta_3$ \\
\hline \hline
\end{tabular}
\label{tableau_robot}
  \end{center}
\end{table}
\subsection{Kinematic modeling}
The velocity $\dot{\bf p}$ of point $P$ can be obtained in three
different forms, depending on which leg is traversed, namely,
\begin{eqnarray}
   {\bf \dot{p}} &=& \dot{\alpha}_1 \negr E (\negr c_1 - \negr a_1) + \dot{\delta}_1 \negr E (\negr c_1 - \negr b_1) +
               \dot{\phi} \negr E (\negr p-\negr c_1) \label{eq:VAMkinematics1}\\
   {\bf \dot{p}} &=& \dot{\alpha}_2 \negr E (\negr c_2 - \negr a_2) +
               \dot{\delta}_2 \negr E (\negr c_2 - \negr b_2) +
               \dot{\phi} \negr E (\negr p-\negr c_2) \label{eq:VAMkinematics2}\\
   {\bf \dot{p}} &=& \dot{\alpha}_3 \negr E (\negr c_3 - \negr a_3) +
               \dot{\delta}_3 \negr E (\negr c_3 - \negr b_3) +
               \dot{\phi} \negr E (\negr p-\negr c_3) \label{eq:VAMkinematics3}
\end{eqnarray}

with matrix ${\bf E}$ defined as
\[
    {\bf E}=\mat{cc}{0&-1\\1&0} \nonumber
\]
$\negr a_i$, $\negr b_i$ and $\negr c_i$ are the position vectors of
points $A_i$, $B_i$ and $C_i$, respectively. $\dot{\alpha}_i$,
$\dot{\delta}_i$ and $\dot{\phi}$ are the rates of angles
$\alpha_i$, $\delta_i$ and $\phi$ depicted in
Fig.~\ref{fig:3RRRvariable}, $i=1,2,3$.

The kinematic model of the VAM under study can be obtained from
Eqs.(\ref{eq:VAMkinematics1})-(c) by eliminating the idle joint
rates. However, the latter depend on the actuating mode of the
mechanism. For instance, $\dot{\delta}_1$, $\dot{\delta}_2$ and
$\dot{\delta}_3$ are idle with the first actuating mode and the
corresponding kinematic model is obtained by dot-multiplying
Eqs.(\ref{eq:VAMkinematics1})-(c) with $(\negr c_i - \negr b_i)^T$,
$i=1,2,3$. Likewise, $\dot{\delta}_1$, $\dot{\delta}_2$ and
$\dot{\alpha}_3$ are idle with the second actuating mode and the
corresponding kinematic model is obtained by dot-multiplying
Eqs.(\ref{eq:VAMkinematics1})-(b) with $(\negr c_i - \negr b_i)^T$,
$i=1,2$, and Eq.(\ref{eq:VAMkinematics3}) with $(\negr c_3 -
\negr a_3)^T$.

The kinematic model of the VAM can now be cast in vector form,
namely, \be
  \negr A \negr t = \negr B  \dot{\bf q} \quad {\rm with } \quad
  \quad \negr t= [{\bf \dot{p}}~\dot{\phi}]^T \quad {\rm and } \quad
  \dot{\bf q} = [\dot{q_1}~\dot{q_2}~\dot{q_3}]^T
\ee with $\dot{\bf q}$ thus being the vector of actuated joint
rates. $\dot{q_i}=\dot{\alpha_i}$ when link $A_iB_i$ is driven and
$\dot{q_i}=\dot{\delta_i}$ when link $A_iE_i$ is driven,
$i=1,2,3$. $\negr A$ and $\negr B$ are respectively, the direct
and the inverse Jacobian matrices of the mechanism, defined as
\begin{eqnarray}
   \negr A &=& \mat{ccc}{
           (\negr c_1 - \negr h_1)^T &  -(\negr c_1 - \negr h_1)^T \negr E (\negr p-\negr c_1) \\
           (\negr c_2 - \negr h_2)^T &  -(\negr c_2 - \negr h_2)^T \negr E (\negr p-\negr c_2) \\
           (\negr c_3 - \negr h_3)^T &  -(\negr c_3 - \negr h_3)^T \negr E (\negr p-\negr c_3)}
           \label{eq:VAMkinematicsA} \\
   \negr B &=& {\rm diag}\left[ (\negr c_i - \negr b_i)^T  \negr E (\negr b_i-\negr a_i)
   \right] , \quad i=1,2,3 \label{eq:VAMkinematicsB}
\end{eqnarray}
where $\negr h_i = \negr b_i$ when link $A_iB_i$ is driven and
$\negr h_i = \negr a_i$ when link $B_iC_i$ is driven,
$i=1,2,3$.

When \negr A is non singular, we obtain the relation
\be \negr t = \negr J \dot{\bf q} \quad \textrm{with} \quad \negr J
= \negr A^{-1}\negr B \label{eq:J} \ee 
Likewise, we obtain
\be \dot{\bf q} = \negr K \negr t \label{eq:K} \ee 
when \negr B is non singular with \negr K denoting the inverse of \negr
J.

\subsection{Singularity analysis}
The singular configurations associated with the direct-kinematic
matrix of PPMs are well known~\cite{Merlet:2006}. For the
3-\underline{R}RR~PPM, such configurations are reached whenever lines $(B_1C_1)$, $(B_2C_2)$ and $(B_3C_3)$ intersect
(possibly at infinity). For the 3-R\underline{R}R~PPM, such
configurations are reached whenever lines
$(A_1C_1)$, $(A_2C_2)$ and $(A_3C_3)$ intersect. Consequently, the
singular configurations associated with the direct-kinematic matrix
of the VAM are reached whenever lines $(H_1C_1)$,
$(H_2C_2)$ and $(H_3C_3)$ intersect where $H_i$ stands for $B_i$ ($A_i$,
resp.) when link $A_iB_i$ ($B_iC_i$, resp.) is driven, $i=1,2,3$.

From Eq.(\ref{eq:VAMkinematicsB}), the singular configurations
associated with the inverse-kinematics of the VAM are reached
whenever points $A_i$, $B_i$, and $C_i$ are aligned.
\subsection{Performance indices}
We focus here on issues pertaining to manipulability or dexterity.
In this regard, we understand these terms in the sense of measures
of distance to singularity, which brings us to the concept of
condition number in~\cite{Golub:89}. Here, we adopt the
\emph{condition number} of the underlying Jacobian matrices based on
the Frobenius norm as a means to quantify distances to singularity and the \emph{transmission angle}.
\subsubsection{Condition number}
The {\em condition number} $\kappa_{F}({\bf M})$ of a $m \times n$
matrix ${\bf M}$, with $m \leq n$, based on the Frobenius norm is defined as follows
\begin{equation}
  \kappa_{F}({\bf M}) = \frac{1}{m} \sqrt{{\rm tr}({\bf M}^T{\bf M}){\rm tr}\left[({\bf M}^T{\bf M})^{-1}\right]}
\end{equation}
Here, the condition number is computed based on the Frobenius norm as the latter produces a condition number that is analytic in terms of the posture parameters whereas the 2-norm does not. Besides, it is much costlier to compute singular values than to compute matrix inverses.

The terms of the direct Jacobian matrix of the VAM are not
homogeneous as they do not have same units. Accordingly, its
condition number is meaningless. Indeed, its singular values cannot
be arranged in order as they are of different nature. However, from~\cite{Li:1990} and \cite{Paden:1988}, the Jacobian can be normalized
by means of a {\em normalizing length}. Later on,
the concept of {\em characteristic length} was introduced in~\cite{Ranjbaran:1995} in order to avoid the random choice of the normalizing
length. For instance, the previous concept was used in~\cite{Chablat:2002} to analyze the kinetostatic performance of manipulators with multiple inverse kinematic solutions, and therefore to select their best {\em working mode}. 


\subsubsection{Transmission angle} The {\em transmission angle} can be used to
assess the quality of force transmission in mechanisms involving
passive joints. Although it is well known and easily computable for
1-DOF or single loop mechanisms~\cite{Balli:2002,Chen:2005}, it is
not extensively used for $n$-DOF mechanical systems ($n>1$)~\cite{Arakelian:2007}.
\begin{figure}
  \centering
  \psfrag{A1}[c][c][0.9]{$A_1$}
  \psfrag{B1}[c][c][0.9]{$B_1$}
  \psfrag{C1}[c][c][0.9]{$C_1$}
  \psfrag{D1}[c][c][0.9]{$E_1$}
  \psfrag{E1}[c][c][0.9]{$D_1$}
  \psfrag{b1}[c][c][0.9]{$\beta_1$}
  \psfrag{g1}[c][c][0.9]{$\gamma_1$}
  \psfrag{p1}[c][c][0.9]{$\psi_1$}
  \psfrag{A2}[c][c][0.9]{$A_2$}
  \psfrag{B2}[c][c][0.9]{$B_2$}
  \psfrag{C2}[c][c][0.9]{$C_2$}
  \psfrag{D2}[c][c][0.9]{$E_2$}
  \psfrag{E2}[c][c][0.9]{$D_2$}
  \psfrag{A3}[c][c][0.9]{$A_3$}
  \psfrag{B3}[c][c][0.9]{$B_3$}
  \psfrag{C3}[c][c][0.9]{$C_3$}
  \psfrag{D3}[c][c][0.9]{$E_3$}
  \psfrag{E3}[c][c][0.9]{$D_3$}
  \psfrag{I1}[c][c][0.9]{$I_1$}
  \psfrag{x}[c][c][0.9]{$x$}
  \psfrag{y}[c][c][0.9]{$y$}
  \psfrag{fci}[c][c][0.9]{$\negr {Fc}_1$}
  \psfrag{vci}[c][c][0.9]{$\negr {Vc}_1$}
  \includegraphics[width=7cm]{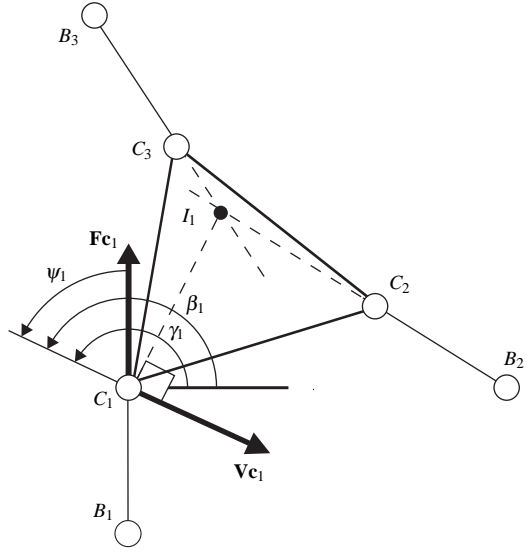}\\
  \caption{Transmission angle of the 3-\underline{R}RR manipulator}
  \label{fig:transmission_angle}
\end{figure}

The transmission angle $\psi_i$ is defined as an angle between vectors
of force $\negr {Fc}_i$ and translational velocity $\negr {Vc}_i$ of
a point to which the force is applied as illustrated in
Fig.~\ref{fig:transmission_angle}. When link $A_iB_i$ is driven, the
direction of force $\negr {Fc}_i$ is the direction of link $B_iC_i$,
namely,
 \be
  \gamma_i= \arctan\left(\frac{y_{C_i}- y_{B_i}}{x_{C_i}- x_{B_i}}\right)  \, , \, i=1,2,3
\ee Conversely, when link $A_iE_i$ is driven, the direction of force
$\negr {Fc}_i$ is the direction of line $(A_iC_i)$, namely, \be
  \gamma_i = \arctan\left(\frac{y_{C_i}- y_{A_i}}{x_{C_i}- x_{A_i}}\right)  \, , \, i=1,2,3
\ee

The instantaneous centre of rotation depends on the leg under study.
For example, instantaneous centre of rotation $I_1$ associated with
leg~1 is the intersecting point of forces $\negr
{Fc}_2$ anf $\negr {Fc}_3$.
\begin{table}
  \begin{center}
  \caption{Cartesian coordinates of instantaneous
centres of rotation}
     \begin{tabular}{|c|c|c|c|c}
     \hline
     & $I_1$ & $I_2$ & $I_3$ \\
     \hline
     $x_{I_i}$ &
     $\displaystyle{\frac{b_3-b_2}{\tan(\gamma_2)-\tan(\gamma_3)}}$ &
     $\displaystyle{\frac{b_1-b_3}{\tan(\gamma_3)-\tan(\gamma_1)}}$ &
     $\displaystyle{\frac{b_2-b_1}{\tan(\gamma_1)-\tan(\gamma_2)}}$ \\ \hline
     $y_{I_i}$ &
     $\displaystyle{\frac{b_3 \tan(\gamma_2)-b_2 \tan(\gamma_3)}{\tan(\gamma_2)-\tan(\gamma_3)}}$ &
     $\displaystyle{\frac{b_1 \tan(\gamma_3)-b_3 \tan(\gamma_1)}{\tan(\gamma_3)-\tan(\gamma_1)}}$ &
     $\displaystyle{\frac{b_2 \tan(\gamma_1)-b_1 \tan(\gamma_2)}{\tan(\gamma_1)-\tan(\gamma_2)}}$ \\ \hline
     \end{tabular}
     \label{table:i}
  \end{center}
\end{table}
Table~\ref{table:i} gives the Cartesian coordinates of instantaneous
centre of rotation $I_i$ associated with the $i$th leg of the VAM,
expressed in frame $\mathcal{F}_b$, with $b_i= y_{C_i} - x_{C_i}
\tan\gamma_i$, $i=1,2,3$.
The direction of $\negr {Vc}_i$ is defined as, \be
  \beta_i= \arctan \left(\frac{y_{C_i}-y_{I_i}}{x_{C_i}-x_{I_i}}\right)+\frac{\pi}{2} \, , \, i=1,2,3
\ee

The transmission angle related to the $i$th leg of the VAM is defined as
follows, \be
  \psi_i= | \gamma_i - \beta_i|  \, , \, i=1,2,3
\ee and the transmission angle $\psi$ of the mechanism is defined as,
\be
  \psi= \max(\psi_i)  \, , \, i=1,2,3 \label{eq:transmissionangle}
\ee

Finally, the smaller $\psi$, the better the force transmission of
the mechanism.
\subsubsection{Regular dextrous workspace} A manipulator had better keep good and homogeneous performance
within its workspace. For that reason, the concept of \emph{regular dextrous workspace} is introduced in~\cite{Chablat:2004}. In
fact, the regular dextrous workspace of a manipulator is a
regular-shaped workspace included in its Cartesian workspace with
good and homogeneous performance. As we focus on the kinetostatic
performance of the VAM in the scope of this paper, we consider only
the condition number of its kinematic Jacobian matrix and its
transmission angle as performance indices.

\section{Actuating Modes Comparison}

For the VAM under study, the inverse condition number of its
kinematic Jacobian matrix, i.e., $\kappa_F^{-1}(\negr J)$ with $\negr
J$ defined in Eq.(\ref{eq:J}), varies from~0 to 1 within its
workspace~$\mathcal{W}$. Likewise, its transmission angle $\psi$,
defined in Eq.(\ref{eq:transmissionangle}), varies from~0 to $90^\circ$
within $\mathcal{W}$. From \cite{Arakelian:2007}, a mechanism
has good kinetostatic performance as long as its transmission angle is
smaller than~$75^\circ$. Let us assume that the kinetostatic performance are good as well as long as $\kappa_F^{-1}(\negr J)>0.15$. Therefore, we claim that the VAM and
its actuating modes~(AMs) have good kinetostatic performance as long
as $\kappa_F^{-1}(\negr J)$ is higher than~0.15 and $\psi$ is smaller
than~$75^\circ$.

First, let us compare the size of the workspace corresponding to AMs
of the VAM given in Table~\ref{tableau_robot}, based on the two
previous kinetostatic performance indices. In this vein, let us
consider that the orientation, $\phi$, of the moving platform of the
VAM is constant and the latter stays as far as possible from
singular configurations, i.e., let $\phi$ be equal to $17.5^\circ$.
From Table~\ref{tab:tabworkspace}, we can notice that the size of
the workspace corresponding to the 2\textsuperscript{nd},
3\textsuperscript{rd} and 4\textsuperscript{th}~AMs is the same.
Likewise, the size of the workspace corresponding to the
4\textsuperscript{th}, 5\textsuperscript{th} and
6\textsuperscript{th}~AMs is the same. This is due to the symmetric
architecture of the mechanism. Moreover, the largest workspace is
obtained with the 1\textsuperscript{st}~AM and the smallest one with
the 8\textsuperscript{th}~AM. Finally, we can notice that the two
kinematic performance indices give similar results.
\begin{table}
  \centering
  \caption{Ratio of the VAM actuating modes workspace size to the VAM workspace size with $\kappa_F^{-1}(\negr J)>0.15$ and $\psi<75^\circ$, $\phi=17.5^\circ$}
       \begin{tabular}{c|c|c}
       \hline \hline
        Actuating mode & \multicolumn{2}{|c}{Workspace size ratio [\%]}  \\
    \cline{2-3}
    number  & $\kappa_F^{-1}(\negr J)>0.15$ & $\psi<75^\circ$ \\
       \hline
       1 & 88.27 & 83.16 \\
       2,3,4 & 75.33 & 71.93 \\
       5,6,7 & 62.26 & 70.76 \\
       8 & 52.15 & 71.86 \\
       \hline \hline
       \end{tabular}
       \label{tab:tabworkspace}
\end{table}

\begin{figure}
\centering
    \subfigure[]
    {
    \psfrag{s01}[t][t][0.7]{$x$}%
    \psfrag{s02}[b][b][0.7]{$y$}%
    %
    \psfrag{x01}[t][t][0.7]{1}%
    \psfrag{x02}[t][t][0.7]{2}%
    \psfrag{x03}[t][t][0.7]{3}%
    \psfrag{x04}[t][t][0.7]{4}%
    \psfrag{x05}[t][t][0.7]{}%
    \psfrag{x06}[t][t][0.7]{6}%
    \psfrag{x07}[t][t][0.7]{7}%
    \psfrag{x08}[t][t][0.7]{8}%
    \psfrag{x09}[t][t][0.7]{9}%
    %
    \psfrag{v01}[r][r][0.7]{0}%
    \psfrag{v02}[r][r][0.7]{1}%
    \psfrag{v03}[r][r][0.7]{2}%
    \psfrag{v04}[r][r][0.7]{3}%
    \psfrag{v05}[r][r][0.7]{4}%
    \psfrag{v06}[r][r][0.7]{5}%
    \psfrag{v07}[r][r][0.7]{6}%
    \psfrag{v08}[r][r][0.7]{7}%
    \includegraphics[height=2.9cm]{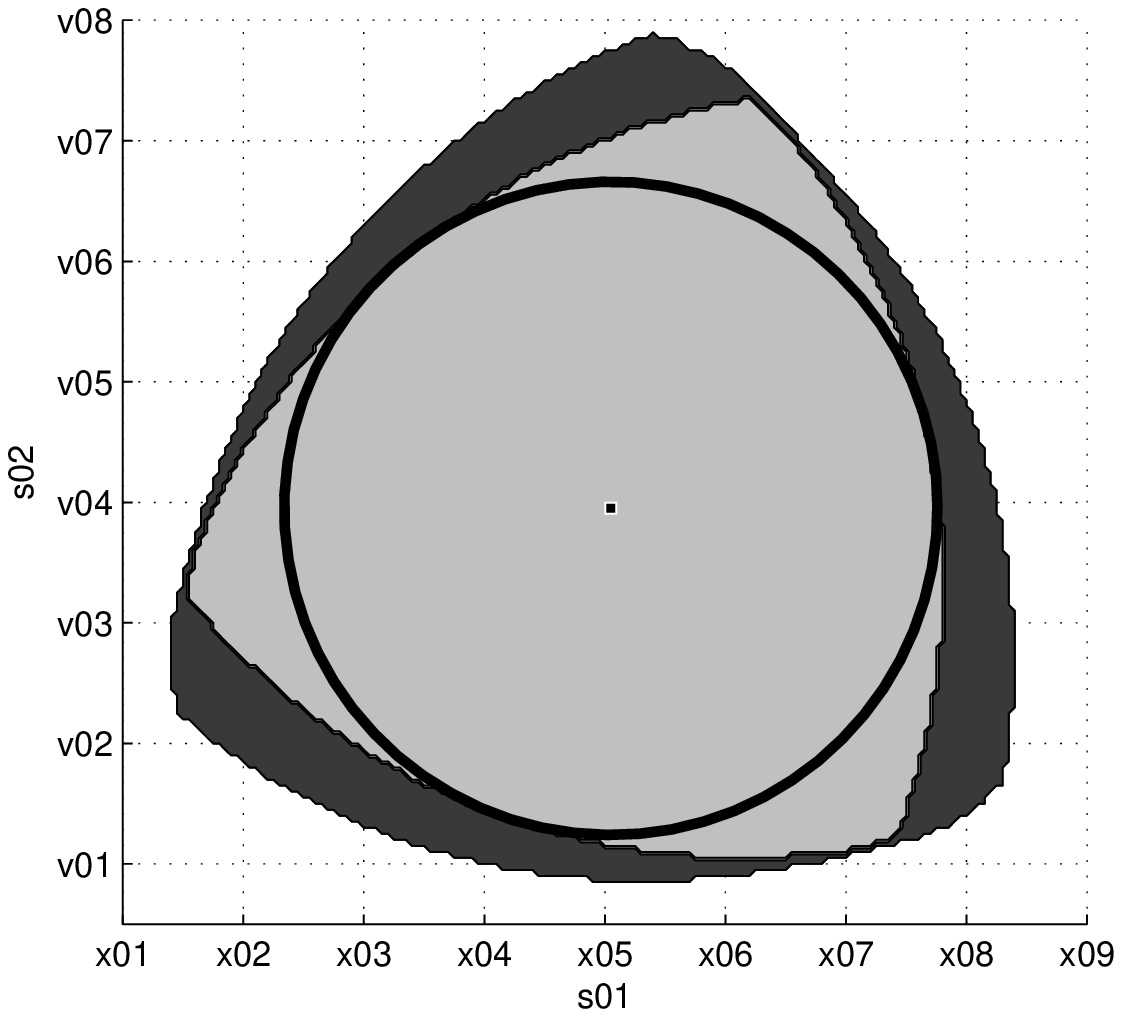}
    }
    \subfigure[]
    {
    \psfrag{s01}[t][t][0.7]{$x$}%
    \psfrag{s02}[b][b][0.7]{$y$}%
    %
    \psfrag{x01}[t][t][0.7]{1}%
    \psfrag{x02}[t][t][0.7]{2}%
    \psfrag{x03}[t][t][0.7]{3}%
    \psfrag{x04}[t][t][0.7]{4}%
    \psfrag{x05}[t][t][0.7]{}%
    \psfrag{x06}[t][t][0.7]{6}%
    \psfrag{x07}[t][t][0.7]{7}%
    \psfrag{x08}[t][t][0.7]{8}%
    \psfrag{x09}[t][t][0.7]{9}%
    %
    \psfrag{v01}[r][r][0.7]{0}%
    \psfrag{v02}[r][r][0.7]{1}%
    \psfrag{v03}[r][r][0.7]{2}%
    \psfrag{v04}[r][r][0.7]{3}%
    \psfrag{v05}[r][r][0.7]{4}%
    \psfrag{v06}[r][r][0.7]{5}%
    \psfrag{v07}[r][r][0.7]{6}%
    \psfrag{v08}[r][r][0.7]{7}%
    \includegraphics[height=2.9cm]{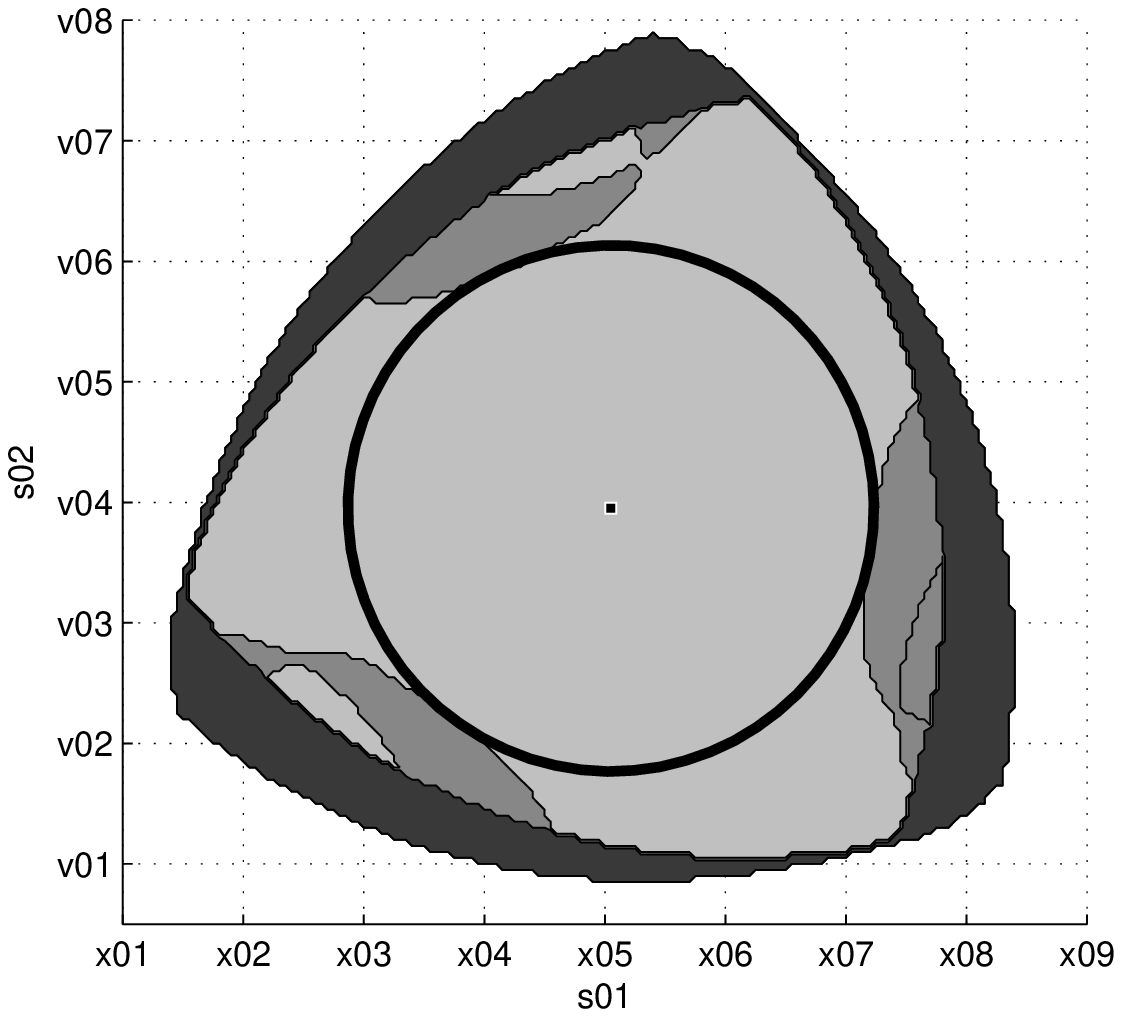}
    }
    \subfigure[]
    {
    \psfrag{s01}[t][t][0.7]{$x$}%
    \psfrag{s02}[b][b][0.7]{$y$}%
    %
    \psfrag{x01}[t][t][0.7]{1}%
    \psfrag{x02}[t][t][0.7]{2}%
    \psfrag{x03}[t][t][0.7]{3}%
    \psfrag{x04}[t][t][0.7]{4}%
    \psfrag{x05}[t][t][0.7]{}%
    \psfrag{x06}[t][t][0.7]{6}%
    \psfrag{x07}[t][t][0.7]{7}%
    \psfrag{x08}[t][t][0.7]{8}%
    \psfrag{x09}[t][t][0.7]{9}%
    %
    \psfrag{v01}[r][r][0.7]{0}%
    \psfrag{v02}[r][r][0.7]{1}%
    \psfrag{v03}[r][r][0.7]{2}%
    \psfrag{v04}[r][r][0.7]{3}%
    \psfrag{v05}[r][r][0.7]{4}%
    \psfrag{v06}[r][r][0.7]{5}%
    \psfrag{v07}[r][r][0.7]{6}%
    \psfrag{v08}[r][r][0.7]{7}%
    \includegraphics[height=2.9cm]{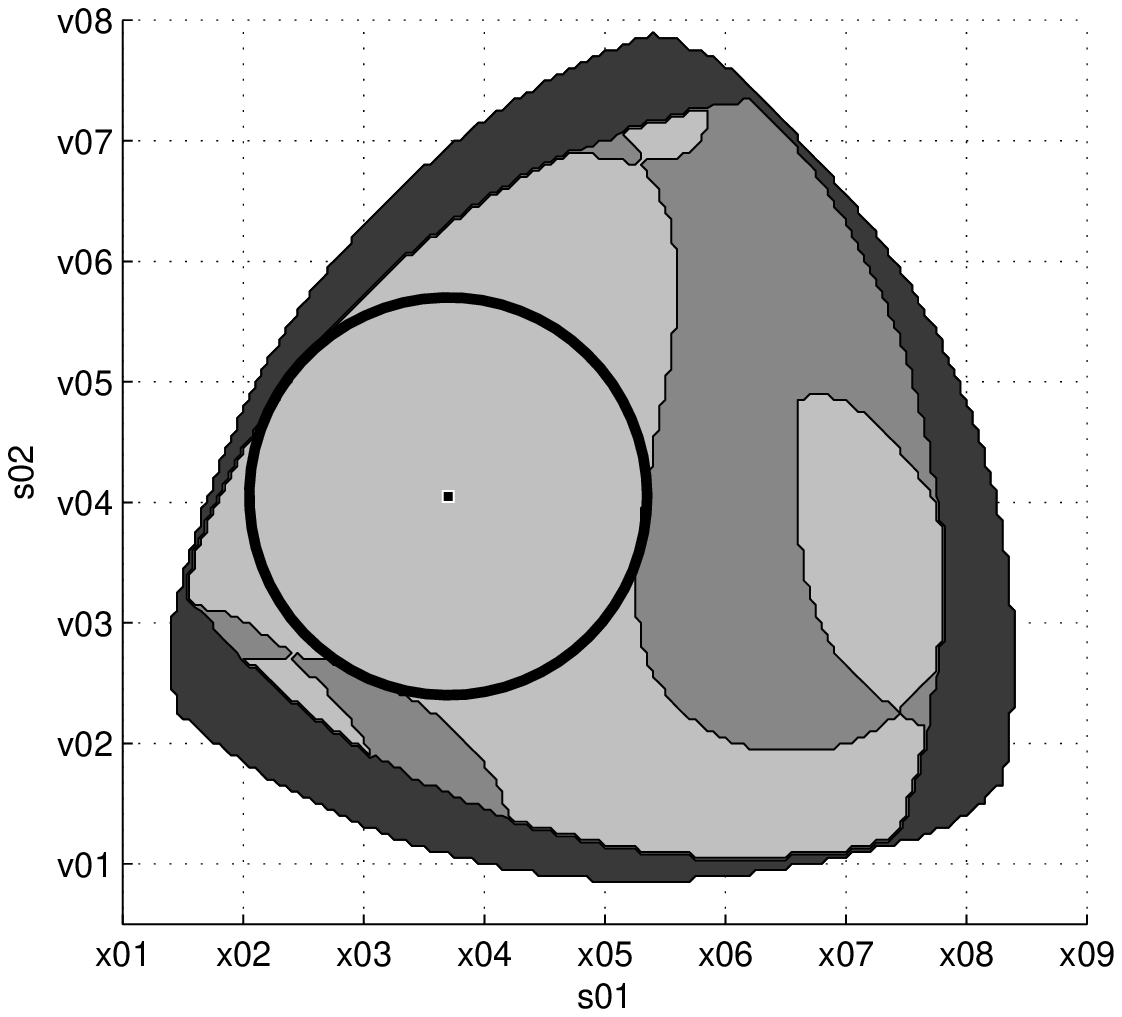}
    }
    \subfigure[]
    {
    \psfrag{s01}[t][t][0.7]{$x$}%
    \psfrag{s02}[b][b][0.7]{$y$}%
    \psfrag{x01}[t][t][0.7]{1}%
    \psfrag{x02}[t][t][0.7]{2}%
    \psfrag{x03}[t][t][0.7]{3}%
    \psfrag{x04}[t][t][0.7]{4}%
    \psfrag{x05}[t][t][0.7]{}%
    \psfrag{x06}[t][t][0.7]{6}%
    \psfrag{x07}[t][t][0.7]{7}%
    \psfrag{x08}[t][t][0.7]{8}%
    \psfrag{x09}[t][t][0.7]{9}%
    %
    \psfrag{v01}[r][r][0.7]{0}%
    \psfrag{v02}[r][r][0.7]{1}%
    \psfrag{v03}[r][r][0.7]{2}%
    \psfrag{v04}[r][r][0.7]{3}%
    \psfrag{v05}[r][r][0.7]{4}%
    \psfrag{v06}[r][r][0.7]{5}%
    \psfrag{v07}[r][r][0.7]{6}%
    \psfrag{v08}[r][r][0.7]{7}%
    \includegraphics[height=2.9cm]{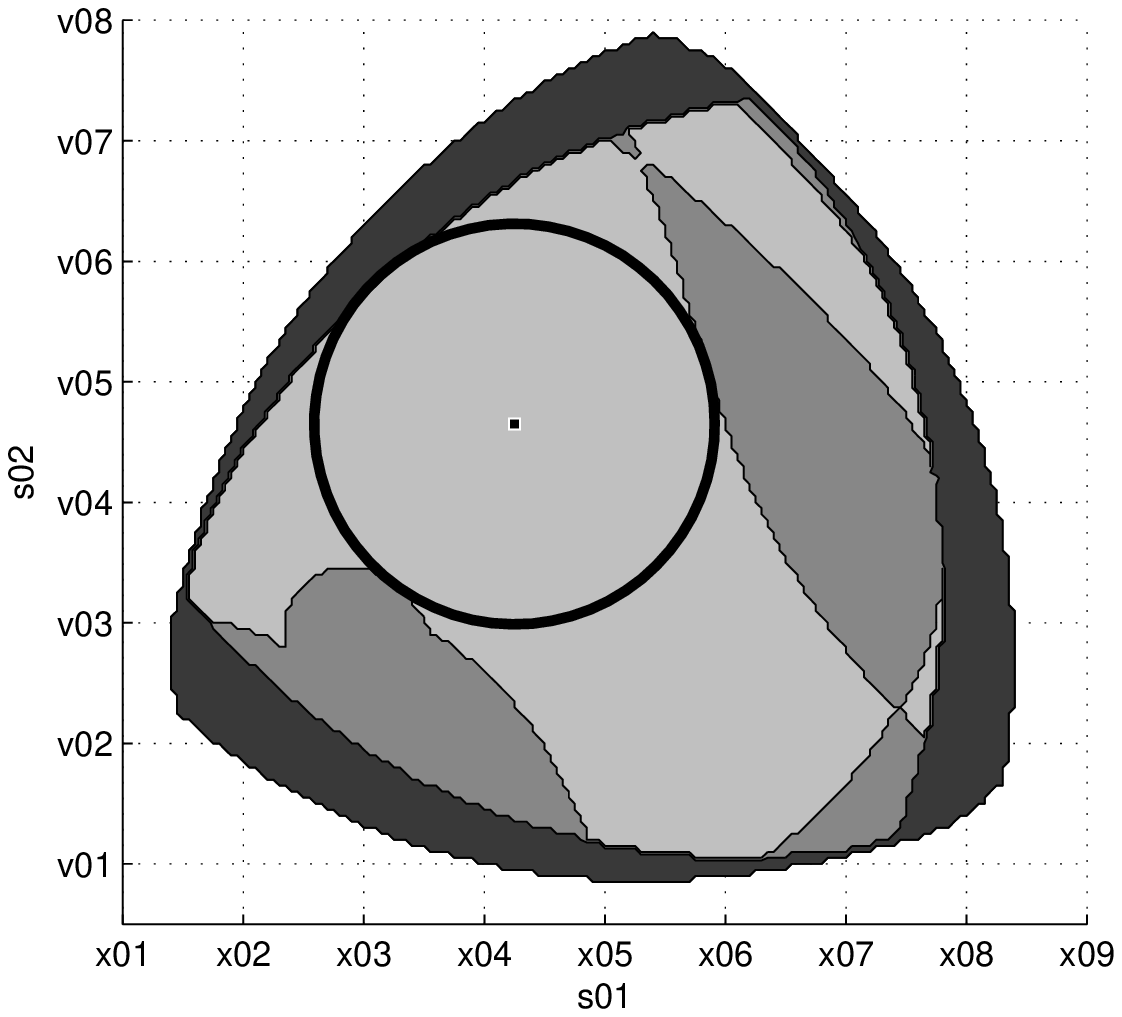}
    }
    \subfigure[]
    {
    \psfrag{s01}[t][t][0.7]{$x$}%
    \psfrag{s02}[b][b][0.7]{$y$}%
    \psfrag{x01}[t][t][0.7]{1}%
    \psfrag{x02}[t][t][0.7]{2}%
    \psfrag{x03}[t][t][0.7]{3}%
    \psfrag{x04}[t][t][0.7]{4}%
    \psfrag{x05}[t][t][0.7]{}%
    \psfrag{x06}[t][t][0.7]{6}%
    \psfrag{x07}[t][t][0.7]{7}%
    \psfrag{x08}[t][t][0.7]{8}%
    \psfrag{x09}[t][t][0.7]{9}%
    %
    \psfrag{v01}[r][r][0.7]{0}%
    \psfrag{v02}[r][r][0.7]{1}%
    \psfrag{v03}[r][r][0.7]{2}%
    \psfrag{v04}[r][r][0.7]{3}%
    \psfrag{v05}[r][r][0.7]{4}%
    \psfrag{v06}[r][r][0.7]{5}%
    \psfrag{v07}[r][r][0.7]{6}%
    \psfrag{v08}[r][r][0.7]{7}%
    \includegraphics[height=2.9cm]{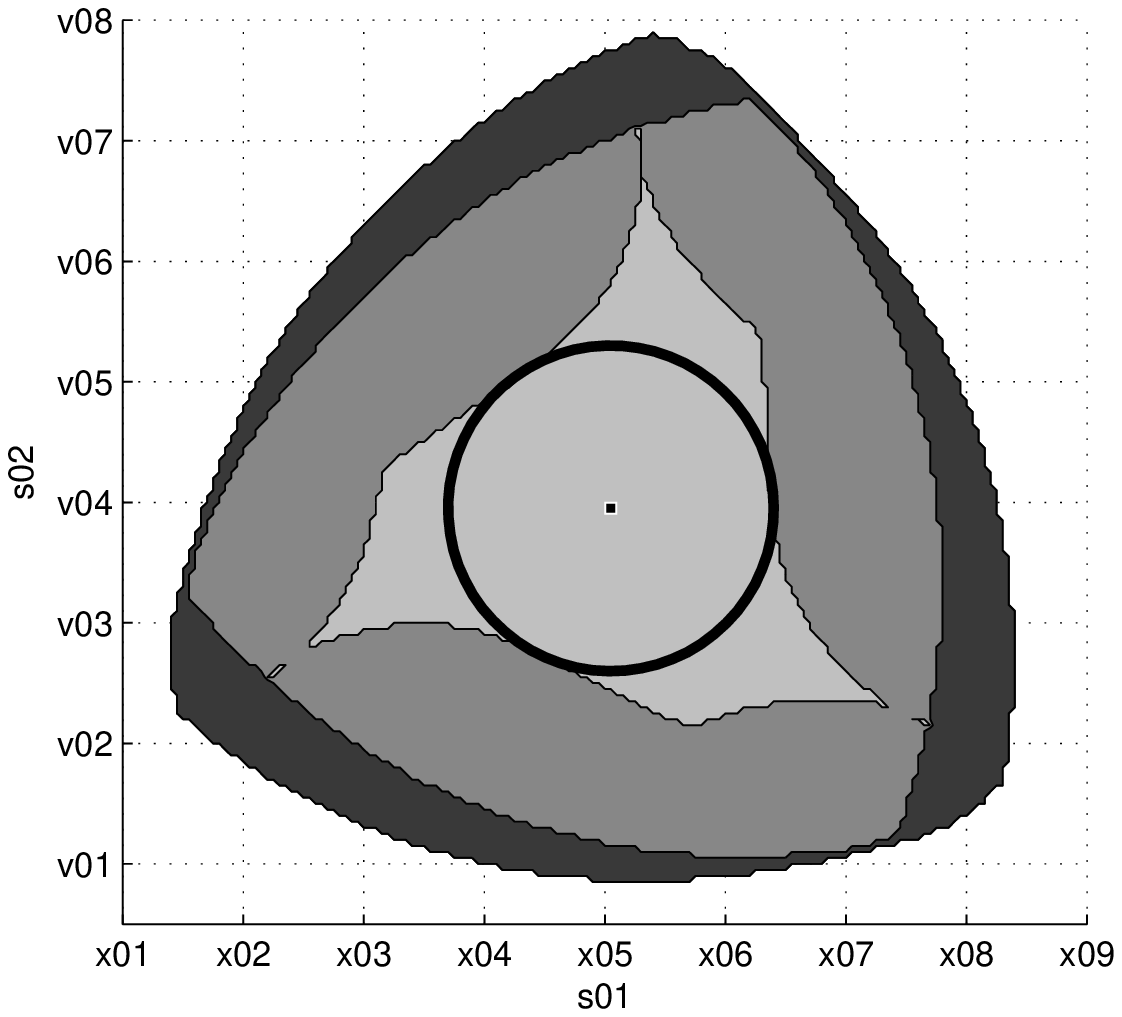}
    }
    \caption{RDW obtained with
$\psi<75^\circ$ of the (a)~VAM; (b)~1\textsuperscript{st} AM;
(c)~2\textsuperscript{nd}, 3\textsuperscript{rd} and
4\textsuperscript{th}~AM; (d)~5\textsuperscript{th},
6\textsuperscript{th} and 7\textsuperscript{th}~AM;
(e)~8\textsuperscript{th} AM} \label{fig:angle_pression_acc}
\end{figure}

In order to compare the AMs of the VAM, we also assume that its
regular dextrous workspace (RDW) is a cylinder, of which the section
depicts the position ($x,y$) of its moving platform and the height
shows the rotation $\phi$ of the latter. Let $\phi$ vary between
$5^\circ$ and $25^\circ$.
\begin{figure}
\centering
    \subfigure[]
    {
    \psfrag{s01}[t][t][0.7]{$x$}%
    \psfrag{s02}[b][b][0.7]{$y$}%
    %
    \psfrag{x01}[t][t][0.7]{1}%
    \psfrag{x02}[t][t][0.7]{2}%
    \psfrag{x03}[t][t][0.7]{3}%
    \psfrag{x04}[t][t][0.7]{4}%
    \psfrag{x05}[t][t][0.7]{}%
    \psfrag{x06}[t][t][0.7]{6}%
    \psfrag{x07}[t][t][0.7]{7}%
    \psfrag{x08}[t][t][0.7]{8}%
    \psfrag{x09}[t][t][0.7]{9}%
    %
    \psfrag{v01}[r][r][0.7]{0}%
    \psfrag{v02}[r][r][0.7]{1}%
    \psfrag{v03}[r][r][0.7]{2}%
    \psfrag{v04}[r][r][0.7]{3}%
    \psfrag{v05}[r][r][0.7]{4}%
    \psfrag{v06}[r][r][0.7]{5}%
    \psfrag{v07}[r][r][0.7]{6}%
    \psfrag{v08}[r][r][0.7]{7}%
    \includegraphics[height=2.9cm]{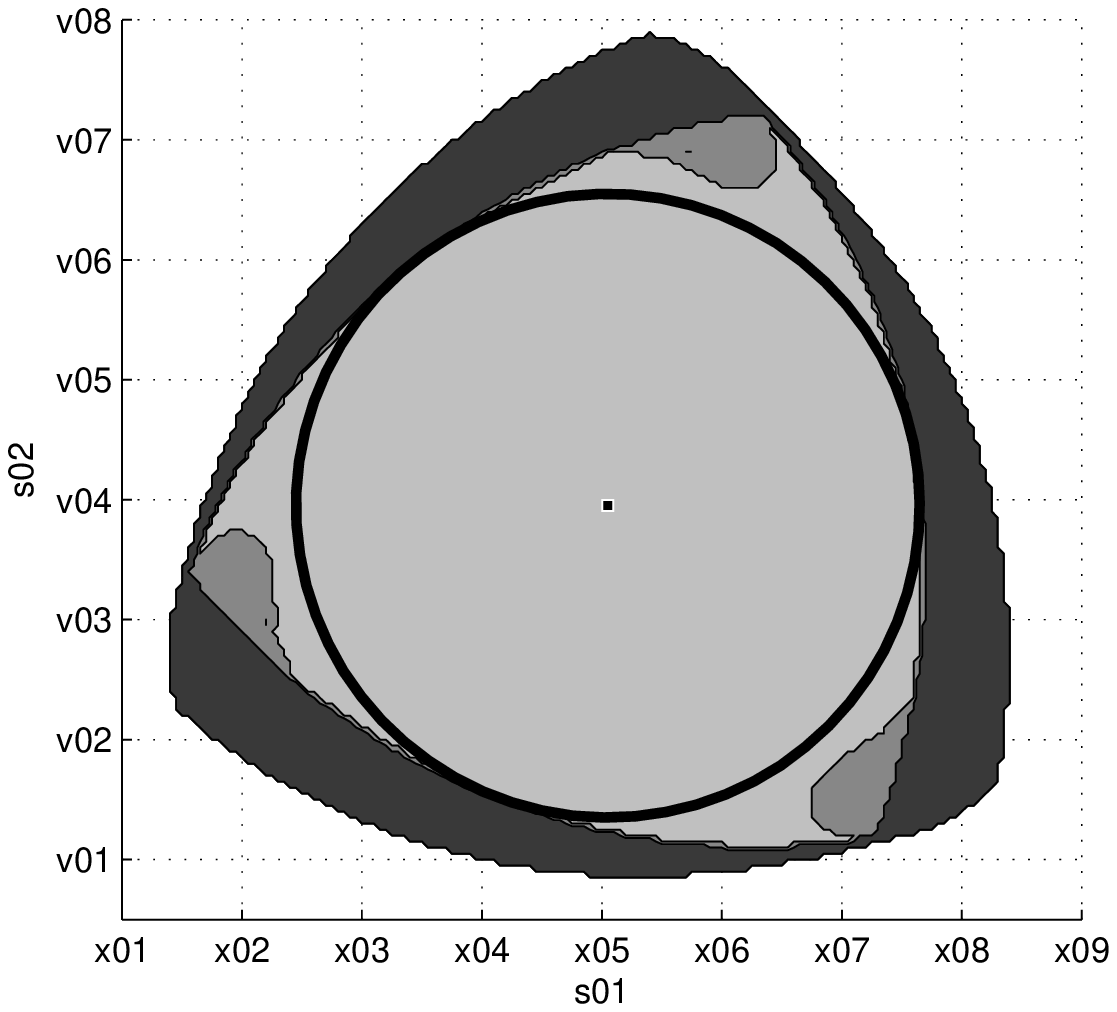}
    }
    \subfigure[]
    {
    \psfrag{s01}[t][t][0.7]{$x$}%
    \psfrag{s02}[b][b][0.7]{$y$}%
    %
    \psfrag{x01}[t][t][0.7]{1}%
    \psfrag{x02}[t][t][0.7]{2}%
    \psfrag{x03}[t][t][0.7]{3}%
    \psfrag{x04}[t][t][0.7]{4}%
    \psfrag{x05}[t][t][0.7]{}%
    \psfrag{x06}[t][t][0.7]{6}%
    \psfrag{x07}[t][t][0.7]{7}%
    \psfrag{x08}[t][t][0.7]{8}%
    \psfrag{x09}[t][t][0.7]{9}%
    %
    \psfrag{v01}[r][r][0.7]{0}%
    \psfrag{v02}[r][r][0.7]{1}%
    \psfrag{v03}[r][r][0.7]{2}%
    \psfrag{v04}[r][r][0.7]{3}%
    \psfrag{v05}[r][r][0.7]{4}%
    \psfrag{v06}[r][r][0.7]{5}%
    \psfrag{v07}[r][r][0.7]{6}%
    \psfrag{v08}[r][r][0.7]{7}%
    \includegraphics[height=2.9cm]{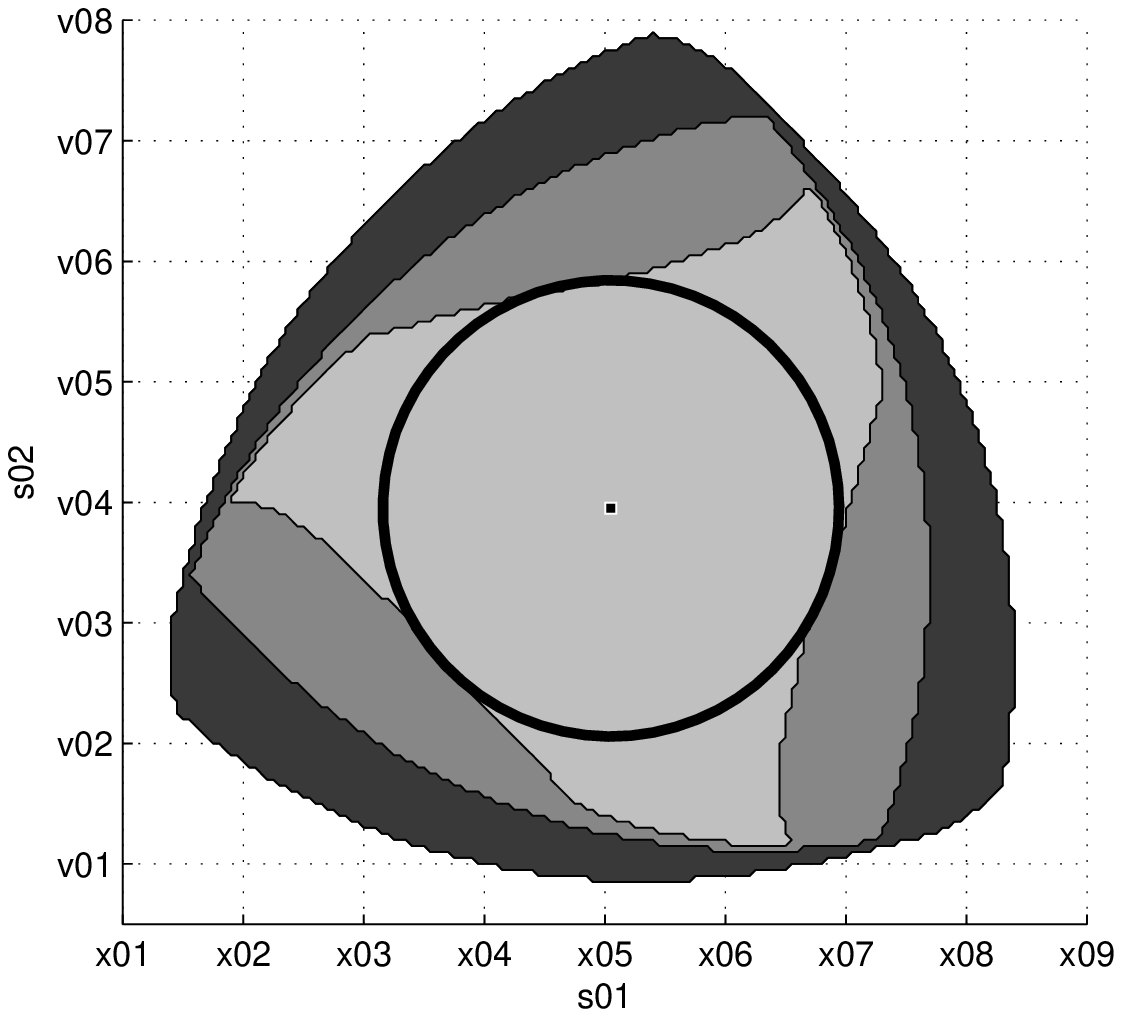}
    }
    \subfigure[]
    {
    \psfrag{s01}[t][t][0.7]{$x$}%
    \psfrag{s02}[b][b][0.7]{$y$}%
    %
    \psfrag{x01}[t][t][0.7]{1}%
    \psfrag{x02}[t][t][0.7]{2}%
    \psfrag{x03}[t][t][0.7]{3}%
    \psfrag{x04}[t][t][0.7]{4}%
    \psfrag{x05}[t][t][0.7]{}%
    \psfrag{x06}[t][t][0.7]{6}%
    \psfrag{x07}[t][t][0.7]{7}%
    \psfrag{x08}[t][t][0.7]{8}%
    \psfrag{x09}[t][t][0.7]{9}%
    %
    \psfrag{v01}[r][r][0.7]{0}%
    \psfrag{v02}[r][r][0.7]{1}%
    \psfrag{v03}[r][r][0.7]{2}%
    \psfrag{v04}[r][r][0.7]{3}%
    \psfrag{v05}[r][r][0.7]{4}%
    \psfrag{v06}[r][r][0.7]{5}%
    \psfrag{v07}[r][r][0.7]{6}%
    \psfrag{v08}[r][r][0.7]{7}
    \includegraphics[height=2.9cm]{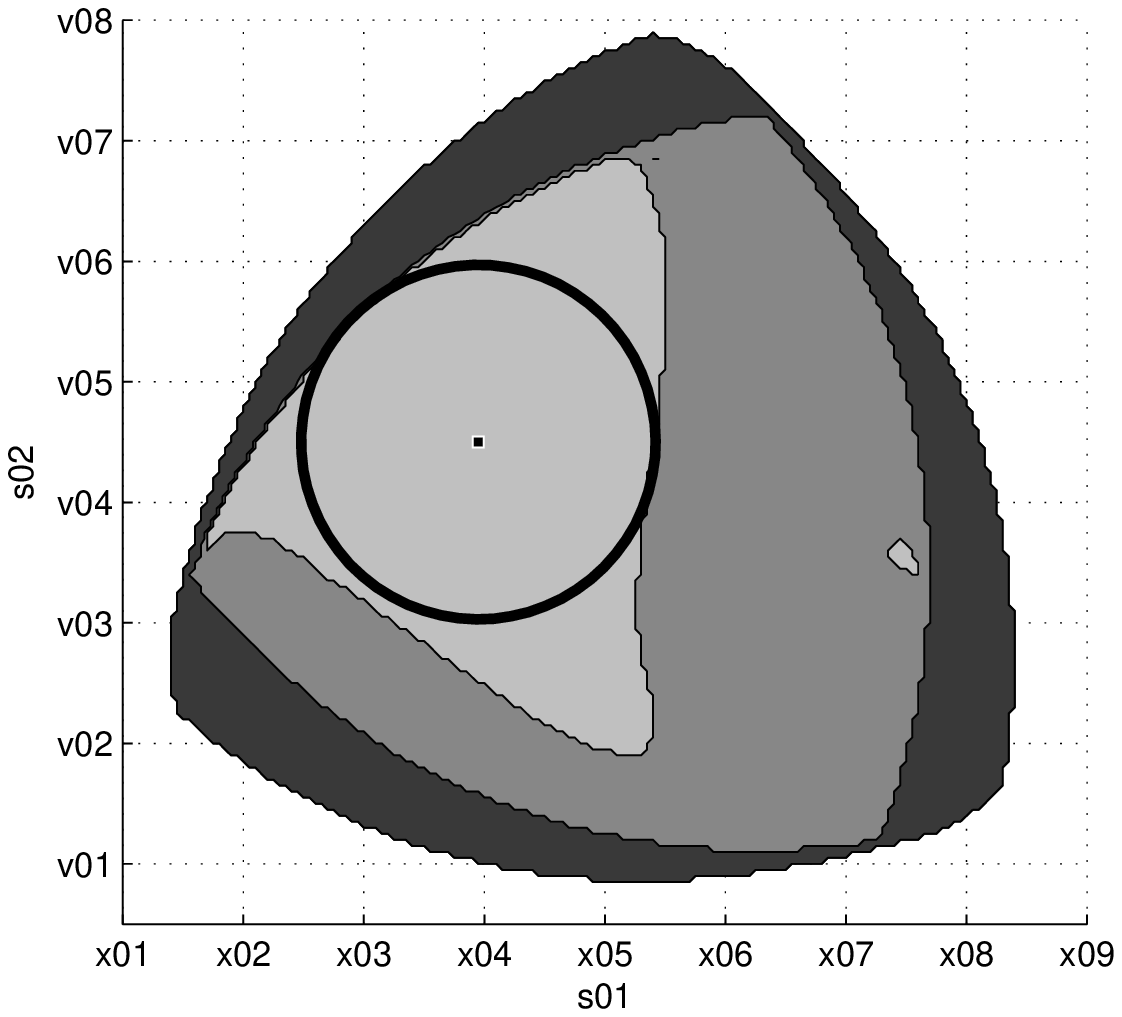}
    }
    \subfigure[]
    {
    \psfrag{s01}[t][t][0.7]{$x$}%
    \psfrag{s02}[b][b][0.7]{$y$}%
    \psfrag{x01}[t][t][0.7]{1}%
    \psfrag{x02}[t][t][0.7]{2}%
    \psfrag{x03}[t][t][0.7]{3}%
    \psfrag{x04}[t][t][0.7]{4}%
    \psfrag{x05}[t][t][0.7]{}%
    \psfrag{x06}[t][t][0.7]{6}%
    \psfrag{x07}[t][t][0.7]{7}%
    \psfrag{x08}[t][t][0.7]{8}%
    \psfrag{x09}[t][t][0.7]{9}%
    %
    \psfrag{v01}[r][r][0.7]{0}%
    \psfrag{v02}[r][r][0.7]{1}%
    \psfrag{v03}[r][r][0.7]{2}%
    \psfrag{v04}[r][r][0.7]{3}%
    \psfrag{v05}[r][r][0.7]{4}%
    \psfrag{v06}[r][r][0.7]{5}%
    \psfrag{v07}[r][r][0.7]{6}%
    \psfrag{v08}[r][r][0.7]{7}%
    \includegraphics[height=2.9cm]{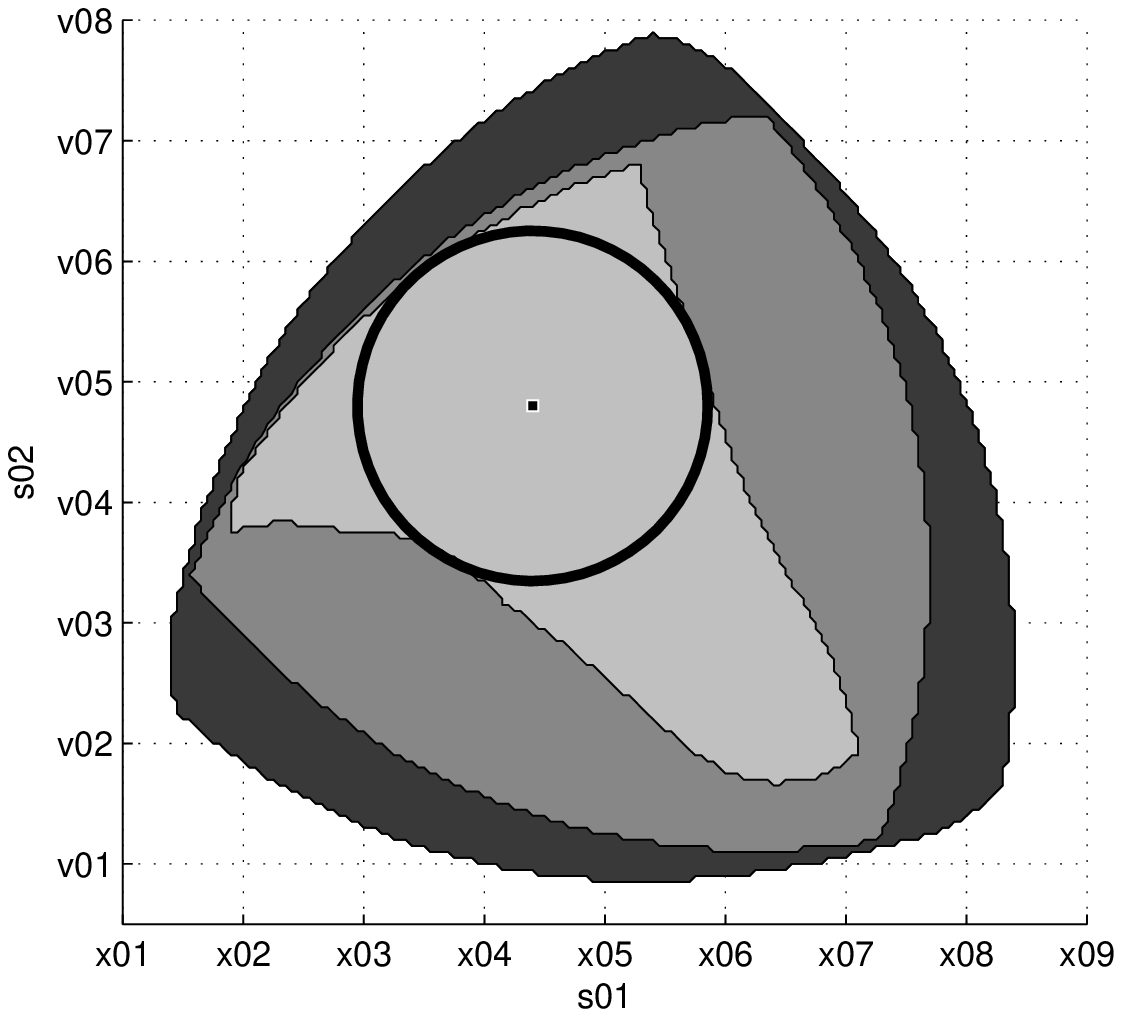}
    }
    \subfigure[]
    {
    \psfrag{s01}[t][t][0.7]{$x$}%
    \psfrag{s02}[b][b][0.7]{$y$}%
    %
    \psfrag{x01}[t][t][0.7]{1}%
    \psfrag{x02}[t][t][0.7]{2}%
    \psfrag{x03}[t][t][0.7]{3}%
    \psfrag{x04}[t][t][0.7]{4}%
    \psfrag{x05}[t][t][0.7]{}%
    \psfrag{x06}[t][t][0.7]{6}%
    \psfrag{x07}[t][t][0.7]{7}%
    \psfrag{x08}[t][t][0.7]{8}%
    \psfrag{x09}[t][t][0.7]{9}%
    %
    \psfrag{v01}[r][r][0.7]{0}%
    \psfrag{v02}[r][r][0.7]{1}%
    \psfrag{v03}[r][r][0.7]{2}%
    \psfrag{v04}[r][r][0.7]{3}%
    \psfrag{v05}[r][r][0.7]{4}%
    \psfrag{v06}[r][r][0.7]{5}%
    \psfrag{v07}[r][r][0.7]{6}%
    \psfrag{v08}[r][r][0.7]{7}%
    \includegraphics[height=2.9cm]{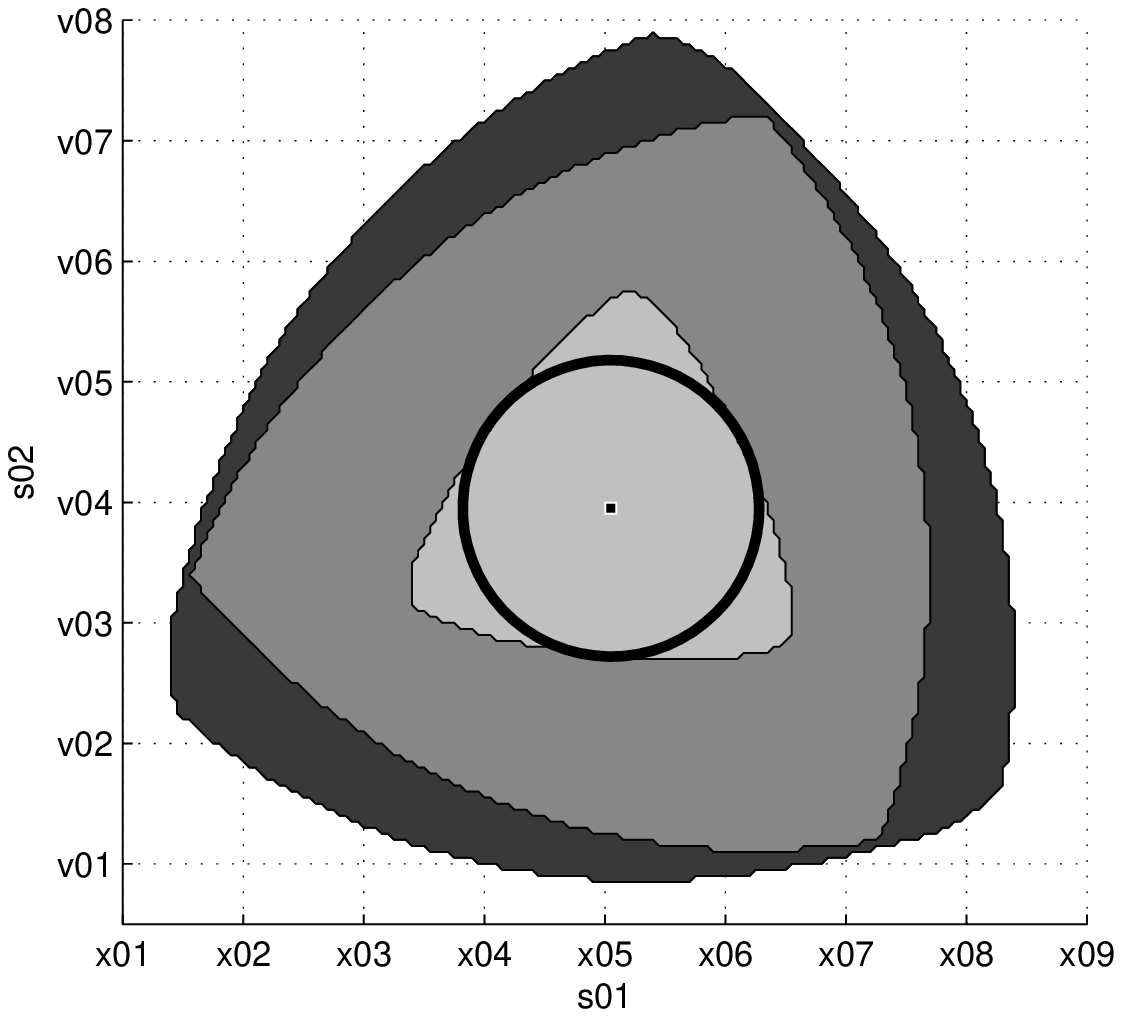}
    }
    \caption{RDW obtained with
$\kappa_F^{-1}(\negr J)>0.15$ of the (a)~VAM;
(b)~1\textsuperscript{st} AM; (c)~2\textsuperscript{nd},
3\textsuperscript{rd} and 4\textsuperscript{th}~AM;
(d)~5\textsuperscript{th}, 6\textsuperscript{th} and
7\textsuperscript{th}~AM; (e)~8\textsuperscript{th} AM}
\label{fig:conditionnement_acc}
\end{figure}
Figures~\ref{fig:conditionnement_acc}(a)-(e)
(Figures~\ref{fig:angle_pression_acc}(a)-(e), resp.) illustrate the
kinetostatic performance of the VAM and its AMs within the workspace
based on $\kappa_F^{-1}(\negr J)$ ($\psi$, resp.). The dark zones
depict the positions of $P$, in which $\phi$ cannot vary
continuously between $5^\circ$ and $25^\circ$. The dark gray zones
depict the positions of $P$, in which $\phi$ can vary continuously
between $5^\circ$ and $25^\circ$, but $\kappa_F^{-1}(\negr J)$
($\psi$, resp.) is not necessarily higher (smaller, resp.) than 0.15
($75^\circ$, resp.). The light gray zones depict the positions of
$P$, in which $\phi$ can vary continuously between $5^\circ$ and
$25^\circ$ and $\kappa_F^{-1}(\negr J)$ ($\psi$, resp.) is higher
(smaller, resp.) than 0.15 ($75^\circ$, resp.). Finally, the circles
describe the RDW of the VAM and its AMs based on $\kappa^{-1}(\negr
J)$ ($\psi$, resp.).

Table~\ref{tab:tabRDWradius} gives RDW radius of the VAM and its AMs
obtained with $\kappa_F^{-1}(\negr J)>0.15$ and $\psi<75^\circ$. We
can notice that the results obtained with the two kinetostatic
performance indices are similar. Besides, the largest RDW is
obtained with the 1\textsuperscript{st}~AM and the smallest one with
8\textsuperscript{th}~AM.

\begin{table}
  \centering
  \caption{RDW radius of the VAM and its AMs obtained with the two kinetostatic performance indices}
       \begin{tabular}{c|c|c}
       \hline \hline
        Actuating mode & \multicolumn{2}{|c}{RDW radius}  \\
    \cline{2-3}
    number  & $\kappa_F^{-1}(\negr J)>0.15$ & $\psi<75^\circ$ \\
       \hline
       1 & 1.89 & 2.18 \\
       2,3,4 & 1.47 & 1.65 \\
       5,6,7 & 1.45 & 1.66 \\
       8 & 1.23 & 1.35 \\
       \hline
       VAM & 2.60 & 2.71 \\
       \hline \hline
       \end{tabular}
       \label{tab:tabRDWradius}
\end{table}

\section{Conclusions}
In this paper, we introduced a new planar parallel mechanism with
variable actuation, which is derived from the architecture of the
3-\underline{R}RR and 3-R\underline{R}R~PPMs. Then, we used two
indices, namely, the condition number of its kinematic Jacobian
matrix and its transmission angle to compare its actuating modes. The
concept of regular dextrous workspace was also used. It turns out
that the mechanism with variable actuation can cover almost all its
workspace with good and homogeneous kinetostatic performance as it
takes advantage of the best performance of its actuating modes.
Finally, for the mechanism at hand, we introduced equivalent bounds
for the condition number and the transmission angle, which allow us to
conclude that the two indices give similar results.

\end{document}